\documentclass[letterpaper]{article}
\usepackage{xcolor}

\usepackage{graphicx}
\usepackage{amssymb}
\usepackage{algorithm}
\usepackage{algorithmicx}
\usepackage{algpseudocode}
\usepackage{subcaption}
\usepackage{graphics}
\usepackage{tabularray}
\usepackage{booktabs}
\usepackage{slashbox}
\usepackage{multirow}
\usepackage{multicol}
\usepackage{array}
\usepackage{adjustbox}
\usepackage{diagbox}
\usepackage{bbm}
\usepackage{makecell}
\usepackage{siunitx}

\usepackage{times}
\usepackage{helvet}
\usepackage{courier}

\usepackage{flairs}

\usepackage{times}
\usepackage{helvet}
\usepackage{courier}

\floatstyle{plain}
\restylefloat{algorithm}

\begin{document}
\title{Lipschitz-Regularized Critics Lead to Policy Robustness Against Transition Dynamics Uncertainty}
\author{Xulin Chen, Ruipeng Liu, Zhenyu Gan, Garrett E. Katz\\
College of Engineering \& Computer Science, Syracuse University\\
Syracuse, NY 13244 U.S.A. \\
\{xchen168, rliu02, zgan02, gkatz01\}@syr.edu
}
\maketitle
\begin{abstract}
\begin{quote}
Uncertainties in transition dynamics pose a critical challenge in reinforcement learning (RL), often resulting in performance degradation of trained policies when deployed on hardware. Many robust RL approaches follow two strategies: enforcing smoothness in actor or actor-critic modules with Lipschitz regularization, or learning robust Bellman operators. However, the first strategy does not investigate the impact of critic-only Lipschitz regularization on policy robustness, while the second lacks comprehensive validation in real-world scenarios. Building on this gap and prior work, we propose PPO-PGDLC, an algorithm based on Proximal Policy Optimization (PPO) that integrates Projected Gradient Descent (PGD) with a Lipschitz-regularized critic (LC). The PGD component calculates the adversarial state within an uncertainty set to approximate the robust Bellman operator, and the Lipschitz-regularized critic further improves the smoothness of learned policies. Experimental results on two classic control tasks and one real-world robotic locomotion task demonstrate that, compared to several baseline algorithms, PPO-PGDLC achieves better performance and predicts smoother actions under environmental perturbations. 
\end{quote}
\end{abstract}

\section{Introduction}

While Reinforcement Learning (RL) has achieved remarkable success in robotic control tasks, its real-world performance on hardware is often hindered by the sim-to-real gap. This gap stems from the mismatch between simulated and real-world transition dynamics, due to system identification errors and idealized modeling. As a result, policies encounter out-of-distribution transitions during hardware deployment and performance degrades. Domain randomization \cite{tobin2017domain,peng2018sim} alleviates this problem by training policies across a range of randomized physical parameters (e.g., mass, friction, and damping) in simulation. However, policies trained by domain randomization may fail to generalize beyond their predefined parameter distributions. This limitation is critical for real-world tasks, where environmental perturbations arise from many sources beyond the predefined domains.

\begin{figure}[t]
    \centering
    \includegraphics{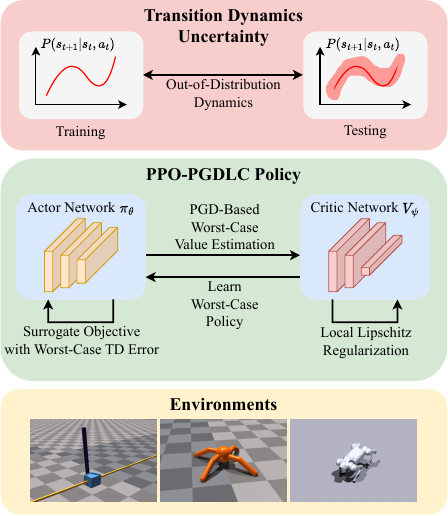}
    \caption{Overview of our work.}
    \label{fig:front-page}
\end{figure}
To generally enhance policy robustness against uncertainties, recent research has turned to robust reinforcement learning (robust RL). One branch of this field addresses transition dynamics uncertainty by optimizing for the worst-case performance within a predefined uncertainty set \cite{iyengar2005robust,abdullah2019wasserstein,wang2021online}. Compared to standard RL, this approach produces conservative but reliable control policies that maintain performance under dynamic mismatches. 
However, directly solving this problem is often intractable for real-world problems due to complex dynamics. State-Conservative Markov Decision Process (SC-MDP) \cite{kuang2022learning} converts the problem into a minimization of the value function over a local region of the state space. Their accompanying solution, called Gradient Based Regularizer (GBR), uses a first-order Taylor expansion to quickly approximate the minimum. Although this approach offers a more tractable design for updating robust RL policies, it still suffers from several limitations. First, the performance of GBR suffers from unregularized gradients in the value network when the local region is relatively large. Furthermore, the efficacy of SC-MDP has not been validated on more complicated tasks such as real-world robotic control problems.

Lipschitz regularization is an effective strategy for enforcing neural network smoothness, with proven success in fields such as computer vision \cite{miyato2018spectral} and RL. In RL, it has been applied to smooth policy outputs against state input perturbations \cite{song2023lipsnet,zhang2024robust} or to stabilize the entire actor-critic module \cite{kobayashi2022l2c2}.
However, the independent effect of a Lipschitz-regularized critic remains underexplored in these previous works. Besides, within the SC-MDP framework, we hypothesize that learning a Lipschitz-regularized critic enforces local smoothness in the value function, therefore reducing the pessimism of the approximated worst-case value function and leading to more robust policy performance. Motivated by SC-MDP and Lipschitz regularization, this work addresses two questions: (1) how sensitive is robust RL policy stability to the method for approximating the value function's local minimum, and (2) do Lipschitz-regularized critic networks enhance policy robustness against transition dynamics uncertainty?

In this work, we propose PPO-PGDLC, a framework shown in Fig. \ref{fig:front-page} that combines PGD-based worst-case value estimation with an LC in the PPO~\cite{schulman2017proximal} framework. 
The PGD component approximates the robust Bellman operator by explicitly searching for adversarial states within a bounded uncertainty set, while the Lipschitz regularization constrains the critic’s gradients to prevent overreaction to local perturbations and to stabilize policy updates. 
Together, these mechanisms produce a principled and computationally tractable formulation of robust RL that connects the theoretical guarantees of SC-MDP with practical actor–critic learning. 
Extensive experiments on classic control benchmarks and real quadrupedal locomotion demonstrate that PPO-PGDLC improves both robustness to transition perturbations and action smoothness relative to existing robust-RL baselines, offering new insight into how critic smoothness contributes to reliable sim-to-real transfer.

\section{Related Work}

\subsection{Robust Reinforcement Learning}

In contrast to traditional RL, which typically assumes a fixed environment, robust RL explicitly accounts for disturbances, uncertainties, and adversarial attacks. These may arise in state space \cite{zhang2020robust,zhang2021robust}, action space \cite{tessler2019action,tan2020robustifying}, rewards \cite{wang2020reinforcement}, or transition dynamics \cite{kuang2022learning,wang2022policy}. When perturbations are properly constrained, techniques such as adversarial training, robust optimization, and game-theoretic approaches can be employed to mitigate their effects \cite{schott2024robust}. Robust RL is relevant both to the reality gap and to unforeseen circumstances that occur during hardware deployment. Such circumstances include unintentional accidents in safety-critical domains such as power grids and autonomous vehicles \cite{ilahi2021challenges}, and intentional deceptions by malicious actors \cite{zhang2021robust,tessler2019action}.

\subsection{Lipschitz Continuity and Regularization}
Modern deep learning systems exhibit sensitivity to small, strategically chosen input perturbations, which can substantially change their outputs \cite{virmaux2018lipschitz,krishnan2020lipschitz}. This vulnerability is also significant in modern RL \cite{song2023lipsnet,pirotta2015policy,kobayashi2022l2c2}, which uses deep learning for policies and value functions. A deep model's Lipschitz constant quantifies its smoothness, and Lipschitz regularization enhances model stability by bounding its Lipschitz constant \cite{yang2020closer,jordan2020exactly}. Lipschitz regularization can be implemented globally, constraining model behavior across the entire input space, or locally, in regions near training data points \cite{jordan2020exactly,shi2022efficiently}. However, Lipschitz regularization presents a fundamental trade-off: excessive regularization causes undue smoothness that impairs performance on the primary task \cite{yang2020closer}.

\section{Preliminaries}

\subsection{Lipschitz Continuity and Regularization}
\label{sec:lipschitz-continuity-and-regularization}
A function $f(\textbf{x}):\mathcal{X} \to \mathcal{Y}$ is considered to be $L$-Lipschitz continuous ($L$-Lipschitz for short) if there is a constant $L\geq0$ such that for all $\textbf{x}_1,\textbf{x}_2 \in \mathcal{X}$, the inequality $d_\mathcal{Y}(f(\textbf{x}_1), f(\textbf{x}_2)) \leq L \cdot d_\mathcal{X}(\textbf{x}_1 ,\textbf{x}_2)$ holds, where $d_{(\cdot)}$ is a distance metric defined on the given space. A lower Lipschitz constant indicates that the function is smoother. The Lipschitz condition also implies that, for all $\textbf{x} \in \mathcal{X}$, the norm of the gradient (if it exists) is upper bounded by $L$, meaning $||\nabla_\textbf{x} f(\textbf{x})|| \leq L$. When $f$ is characterized by a neural network, this property can be used to enforce local smoothness by augmenting the original loss function $\mathcal{L}^\text{ori}$ with a penalty term on the gradient norm, i.e.:
\begin{equation}
    \mathcal{L}^\text{new}(f) = \mathcal{L}^\text{ori}(f) + \lambda \cdot ||\nabla_\textbf{x} f(\textbf{x})||.
    \label{eq:lipschitz-reg-loss}
\end{equation}

\subsection{Robust Markov Decision Process}

A Robust Markov Decision Process (RMDP) \cite{wiesemann2013robust} extends the standard MDP by allowing the transition kernel $P$ to vary within an uncertainty set $\mathcal{P}$ ($P \in \mathcal{P}$). An RMDP is formally described by the tuple $(\mathcal{S}, \mathcal{A}, \mathcal{P}, R, \gamma)$, where $\mathcal{S} \subseteq \mathbb{R}^m$ and $\mathcal{A} \subseteq \mathbb{R}^n$ are the state and action spaces, $\mathcal{P}$ represents the set of possible transition dynamics, $R: \mathcal{S} \times \mathcal{A} \to \mathbb{R}$ is the reward function, and $\gamma \in [0,1)$ is the discount factor for future rewards. Given a distribution $\mu(s_0)$ for the initial state $s_0$, the objective is to find a policy $\pi:\mathcal{S} \to \Delta_{\mathcal{A}}$ which maximizes the infimum expected return across all transition dynamics in $\mathcal{P}$: $ \mathcal{J}_{\mathcal{P}}(\pi) \triangleq \inf_{\hat{P} \in \mathcal{P}} \mathbb{E}_{\mu, \pi, \hat{P}} \left[ \sum_{t=0}^{\infty} \gamma^t R(s_t,a_t) \right]$, where $\Delta_{(\cdot)}$ denotes the set of probability distributions over the given space. For a fixed policy $\pi$, the state- and  action-value functions are $V_{\mathcal{P}}^\pi(s)=\inf_{\hat{P} \in \mathcal{P}} \mathbb{E}_{\pi,\hat{P}} \left[ \sum_{t=0}^{\infty} \gamma^t R(s_t,a_t) |s_0=s \right]$ and $Q_{\mathcal{P}}^\pi(s,a)=\inf_{\hat{P} \in \mathcal{P}} \mathbb{E}_{\pi,\hat{P}} \left[ \sum_{t=0}^{\infty} \gamma^t R(s_t,a_t) |s_0=s,a_0=a \right]$, with respect to the worst-case transition dynamics. Here $Q^\pi_{\mathcal{P}}(s,a)$ is the fixed point of the robust Bellman operator
\begin{equation}
    \mathcal{T}_{\mathcal{P}}^\pi Q(s,a) 
    \triangleq R(s,a) + \gamma \inf_{\hat{P} \in\mathcal{P}} \mathbb{E}_{\substack{\hat{s}\sim \hat{P}(\cdot|s,a)\\ \hat{a}\sim \pi(\cdot|\hat{s})}} \left[ Q(\hat{s},\hat{a}) \right]. \label{eq:bellman-operator-rmdp}
\end{equation}
To make the problem well-posed, one must impose appropriate constraints on the uncertainty set $\mathcal{P}$. A widely used approach \cite{abdullah2019wasserstein,kuang2022learning} is defining a $p^\text{th}$-order Wasserstein $\epsilon$-ball $\mathcal{P}_{\epsilon,p}$ centered at a given nominal dynamics $P_0(\cdot|s,a)$.
In summary, RMDPs enhance policy robustness at the expense of average-case performance.

\section{Methodology}

\subsection{Problem: Worst-Case Value Estimation}

Directly solving (\ref{eq:bellman-operator-rmdp}) is often infeasible for real-world problems, due to complicated dynamics and the difficulty of accurately modeling uncertainties such as variations in weights, friction, or terrain. SC-MDP \cite{kuang2022learning} converts the problem into disturbances within the state space, yielding the modified robust Bellman operator
\begin{equation}
\mathcal{T}_{\mathcal{P}}^\pi Q(s,a) 
    \triangleq R(s,a) + \gamma \inf_{\hat{s} \in B_{\epsilon,p}(s')} \mathbb{E}_{\hat{a} \sim \pi(\cdot|\hat{s})} \left[ Q(\hat{s},\hat{a}) \right], \label{eq:state-conservative-mdp}
\end{equation}
where the state transition $(s,a,s')$ is performed under the nominal dynamics $P_0$, and $B_{\epsilon,p}(s') \triangleq \{\hat{s} \in \mathcal{S}: d_\mathcal{S}(\hat{s}-s')^p \leq \epsilon\}$ is an $\epsilon$-ball induced by a distance metric $d_\mathcal{S}$ on the state space $\mathcal{S}$. This formulation is equivalent to (\ref{eq:bellman-operator-rmdp}) if the dynamics in the uncertainty set $P_{\epsilon,p}$ are deterministic and bounded by Wasserstein distance. Although the equivalence may not hold for stochastic dynamics \cite{kuang2022learning}, (\ref{eq:state-conservative-mdp}) still simplifies the update rule for robust RL policies by directly selecting the most adversarial state transition and eliminating the need for modeling transition dynamics uncertainty. Even if the magnitude of real perturbations exceeds $\epsilon$, policies obtained via learning (\ref{eq:state-conservative-mdp}) exhibit more conservative behavior than standard RL policies.

Actor-critic algorithms in standard RL approximate either the state-value $V(s)$ or the action-value $Q(s,a)$. To adapt these algorithms for robust RL, critic networks must approximate the robust Bellman operator. Our work adapts the on-policy algorithm PPO \cite{schulman2017proximal} to this setting by approximating (\ref{eq:state-conservative-mdp}). Choosing the $\infty$-norm as the distance metric for $B_{\epsilon,p}(s)$, we formalize the problem as Worst-Case Value Estimation (WCVE). Specifically, given a state-value function $V(s)$, a solution $h(s,\epsilon,V)$ of WCVE is a state within a $L_\infty$-norm ball $B_{\epsilon,\infty}(s)$ around the input state $s$, that corresponds to the infimum state-value:
\begin{gather}
    h(s,\epsilon,V) = \arg_{\hat{s} \in B_{\epsilon,\infty}(s)} \inf V(\hat{s}).\label{eq:wcve}
\end{gather}

\subsection{PGD and Critic Lipschitz Regularization}
\label{sec:theoretical-consideration}
Gradient Based Regularizer \cite{kuang2022learning} does not solve $h(s,\epsilon,V)$ in (\ref{eq:wcve}) exactly, but uses the fast approximation $\inf_{\hat{s} \in B_{\epsilon,\infty}(s)}  V(\hat{s}) \approx V(s) - \epsilon ||\nabla_s V(s)||_1$. However, the learned critic network is typically nonconvex, so the approximation is not always reliable. We choose to solve WCVE using PGD \cite{madry2017towards}, an iterative optimization algorithm that searches for solutions within the constraint set. In each iteration, PGD updates the state in the direction that decreases the state-value, then projects the state back into the $L_\infty$-norm ball. Algorithm \ref{alg:wcve-pgd} presents the pseudocode for solving WCVE with PGD.  While PGD does introduce additional overhead, we limit the number of iterations to 10 so that the overhead is not prohibitive.

PGD does not mitigate performance degradation when the gradients of $V$'s deep function approximator are highly inaccurate. To address this limitation, a Lipschitz-regularized critic is considered, which is justified if the true RMDP value functions are Lipschitz. Prior work~\cite{pirotta2015policy,rachelson2010locality} has shown that when the reward function, policy, and transition dynamics satisfy Lipschitz continuity, the resulting state-value and state–action value functions are also Lipschitz continuous. 

In many continuous control tasks, the reward can be designed to be Lipschitz w.r.t. the state-action space. However, dynamics are often not globally Lipschitz because of impulse events that produce discontinuous state changes. This can be mitigated by either designing compliant models for impulse events \cite{castro2020transition} or considering local Lipschitz continuity in dynamical systems \cite{khajenejad2022interval}. We choose the second idea and apply local Lipschitz regularization to the critic network, as in (\ref{eq:lipschitz-reg-loss}).

\begin{algorithm}[t]
\caption{PGD solver $h(s,\epsilon,V)$ for WCVE}\label{alg:wcve-pgd}
\begin{algorithmic}
\Require State-value function $V$, input state $s$, number of PGD steps $N$, PGD step size $\alpha$, constraint set $B_{\epsilon,\infty}(s)$.
\Ensure State with the minimum state-value in $B_{\epsilon,\infty}(s)$.
\State $s^0 \leftarrow s$
\For{$i=0 \dots N-1$}
\State $s^{i+1} \leftarrow s^i - \alpha \cdot \text{sign}(\nabla_s V(s^i))$ 
\State $s^{i+1} = \text{Projection}(s^{i+1}, B_{\epsilon,\infty}(s))$
\EndFor
\State Return $h(s,\epsilon,V) = s^N$.
\end{algorithmic}
\end{algorithm}


\subsection{Practical Implementation}

This section introduces the implementation of PPO-PGDLC, which incorporates the PGD solver and a Lipschitz-regularized critic into PPO. Recall that PPO uses truncated generalized advantage estimation \cite{Schulmanetal_ICLR2016} to reduced variance. Given a trajectory $\tau=(s_0,a_0,s_1,\cdots,s_T)$, the advantage estimator $A_t$ of time index $t\in[0,T)$ is defined as the exponentially weighted sum of temporal difference (TD) errors
\begin{gather}
A_t = \delta_t + (\gamma\xi)\delta_{t+1}+\cdots+(\gamma\xi)^{T-t+1}\delta_{T-1}, \nonumber \\ 
\text{with} \ \delta_t = R(s_t,a_t)+\gamma V_\psi(s_{t+1}) - V_\psi(s_t),
\end{gather}
where $\gamma,\xi  \in (0,1)$ are the discount factor for the future reward and generalized advantage estimation, respectively, and $\delta_t$ is the TD error at time $t$ computed using the critic network $V_\psi$. In PPO-PGDLC, we introduce a modified advantage estimator $\hat{A}^h_t$, which calculates the worst-case TD errors $\hat{\delta}^h_t$ based on the PGD solver $h$ in Algorithm \ref{alg:wcve-pgd}:
\begin{gather}
\hat{A}^h_t= \hat{\delta}^h_t + (\gamma\xi)\hat{\delta}^h_{t+1}+\cdots+(\gamma\xi)^{T-t+1}\hat{\delta}^h_{T-1}, \label{eq:worst-case-gae} \\ 
\text{ with} \ \ \hat{\delta}^h_t = R(s_t,a_t)+\gamma V_\psi(h(s_{t+1},\epsilon,V_\psi)) - V_\psi(s_t). \nonumber
\end{gather}
The actor loss $\mathcal{L}(\theta)$ for policy $\pi_\theta$ is formulated by incorporating $\hat{A}^h_t$ into the PPO surrogate objective, with a clipping ratio $\eta \in (0,1)$:
\begin{align}
    \mathcal{L}(\theta) &= \mathbb{E}_{(s_t,a_t)\sim \mathcal{D}} \Big[ \min \Big( \frac{\pi_{\theta}(a_t|s_t)}{\pi_{\theta_\text{old}}(a_t|s_t)} \hat{A}^h_t, \nonumber \\ 
    & \text{clip}\big( \frac{\pi_{\theta}(a_t|s_t)}{\pi_{\theta_\text{old}}(a_t|s_t)}, 1-\eta,1+\eta  \big) \hat{A}^h_t \Big) \Big].
\end{align} 
The critic loss $\mathcal{L}(\psi)$ consists of a residual error of fitting the $n$-step rewards-to-go $\bar{R}_t=\gamma^n V_\psi (s_{t+n}) + \sum_{i=0}^{n-1} \gamma^t R(s_{t+i},a_{t+i})$, and a local Lipschitz regularization term which controls the strength of smoothness enforcement on the critic as in (\ref{eq:lipschitz-reg-loss}):
\begin{equation}
\mathcal{L}(\psi) = \mathbb{E}_{s_t \sim \mathcal{D}} \left[ \frac{1}{2} \left( V_\psi(s_t) - \bar{R}_t \right)^2 + \lambda ||\nabla_s V_\psi(s_t)||^2_1\right]
\end{equation}
Algorithm \ref{alg:wcve-ppo} presents the complete pseudocode of PPO-PGDLC.

\begin{algorithm}[t]
\caption{PPO-PGDLC}\label{alg:wcve-ppo}
\begin{algorithmic}
\Require Initial actor and critic parameters $\theta,\psi$, rollout buffer $\mathcal{D}$, PGD solver $h$, learning rate $\beta_\theta,\beta_\psi$.
\Ensure Update $\theta$ and $\psi$.
\For{every iteration}
\State Collect a set of trajectories $\{\tau_1,\cdots\tau_n\}$ running $\pi_\theta$, and add them to $\mathcal{D}$.
\State For all transitions in $\mathcal{D}$, compute rewards-to-go $\bar{R}_t$ and advantage estimator $\hat{A}^h_t$ with $V_\psi$ and $h$.
\For{every gradient step}
\State $\theta \leftarrow \theta - \beta_\theta \nabla_\theta \mathcal{L}(\theta)$
\State $\psi \leftarrow \psi - \beta_\psi \nabla_\psi \mathcal{L}(\psi)$
\EndFor
\EndFor
\end{algorithmic}
\end{algorithm}


\section{Experiments}

Experiments are conducted on three control problems: two classic benchmarks, Cartpole and Ant, in Isaac Gym \cite{makoviychuk2021isaac}, and a quadrupedal robot locomotion task using the Unitree Go2, which was trained in simulation and deployed on real hardware \cite{unitreego2}. The experiments primarily aim to compare policy robustness under varying transition dynamics and to evaluate policy smoothness and performance.

\begin{figure*}[t]
    \centering
    \begin{subfigure}[t]{0.2\textwidth}
        \centering
        \includegraphics[width=\textwidth]{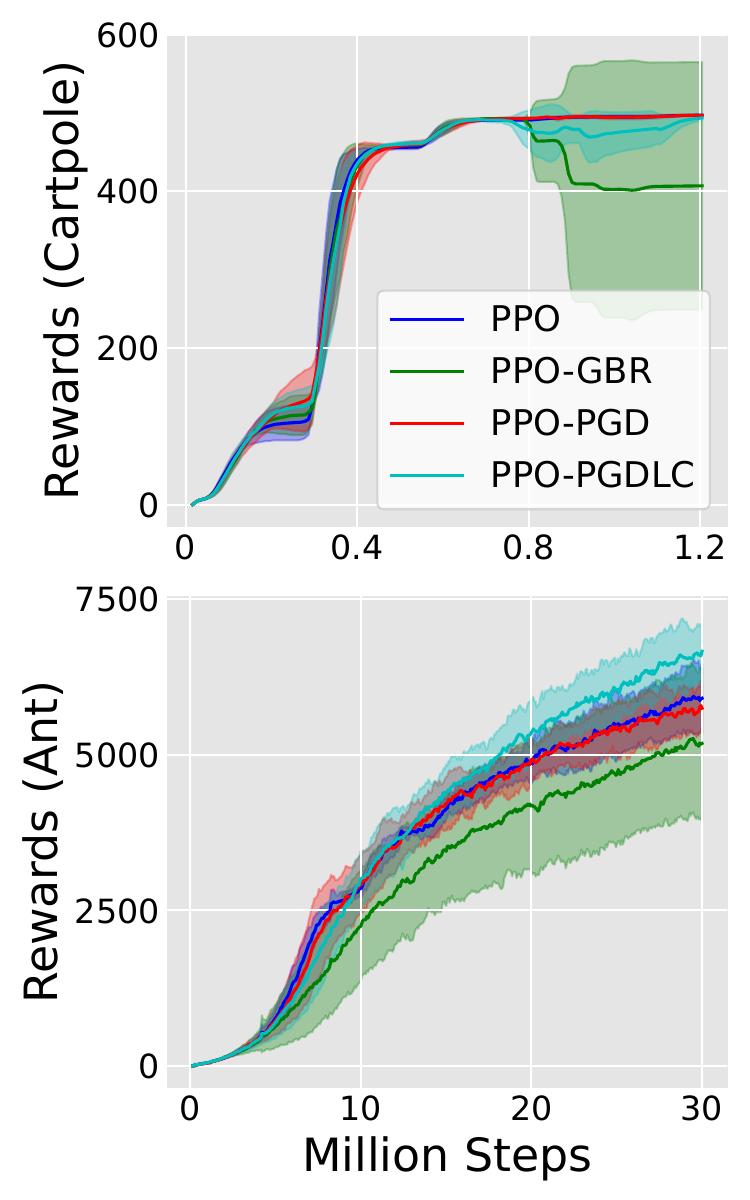}
        \alt{Training reward curve on Cartpole and Ant environment.}
    \end{subfigure}%
    \hspace{0.5cm}
    \begin{subfigure}[t]{0.73\textwidth}
        \centering
        \includegraphics[width=\textwidth]{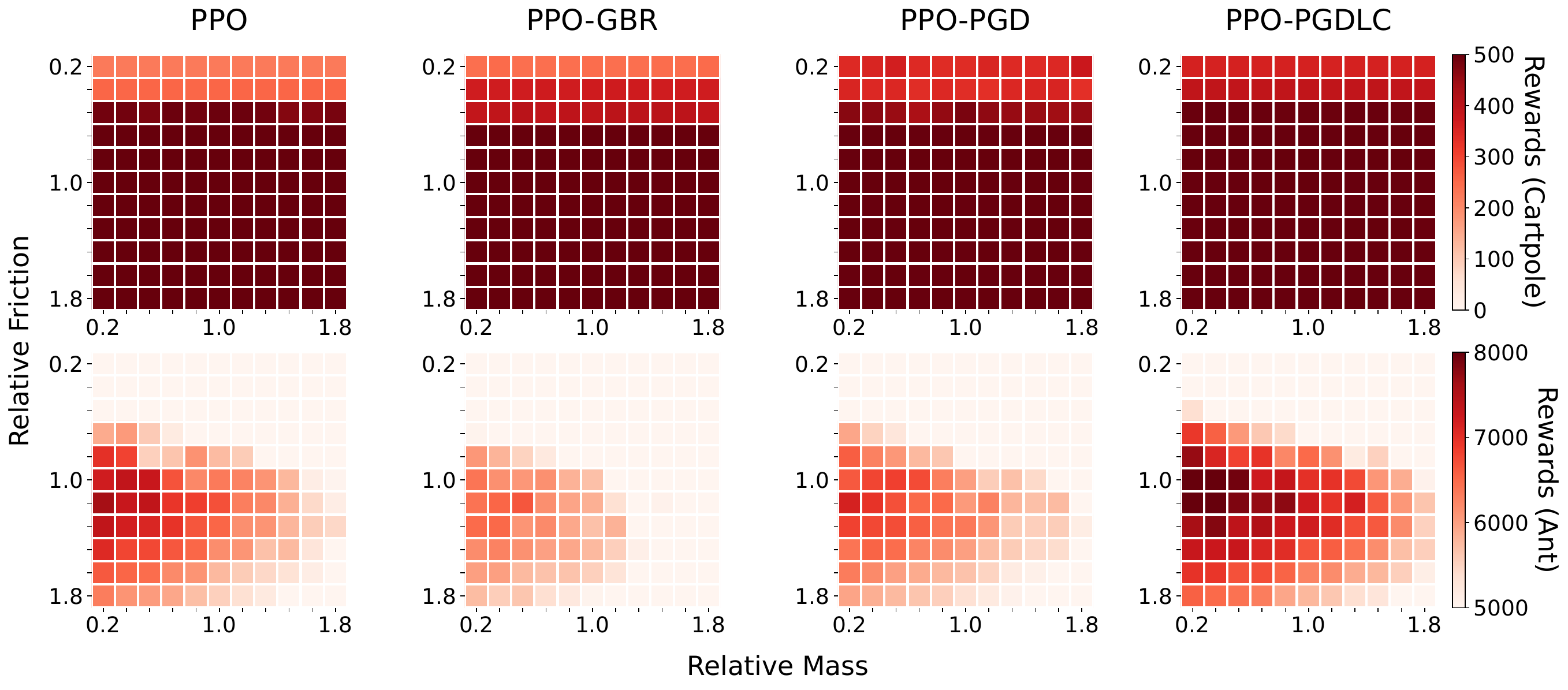}
        \alt{Performance heatmap on varied mass and friction perturbation.}
    \end{subfigure}
    \caption{The training curves (left) and heatmaps (right) for Cartpole (top row) and Ant (bottom row). We evaluate policy robustness against mass and friction perturbation and visualize the average episode rewards as heatmaps. Note that $\epsilon=0.003$ for PPO-GBR, PPO-PGD and PPO-PGDLC. Overall, PPO-PGDLC maintains higher rewards on both the nominal and perturbed environments compared with all baselines.}
    \label{fig:cartpole-ant-results}
\end{figure*}

\subsection{Robustness over Dynamics Perturbation}

We quantify the robustness of learned policies against transition dynamics uncertainty as follows. We generate an $M\times M$ grid of perturbed environments $\mathcal{P}= \{P_{i,j}|1 \leq i,j \leq M\}$ by varying two physical parameters, such as body mass and surface friction. We then design a metric called $\rho$-robustness to capture worst-case performance within a specific ``attack radius'' $\rho$ on the perturbation grid, measured by the $L_\infty$-norm distance from the nominal environment at the grid center. Given a policy $\pi$ and $\rho \in \{0,1,\cdots, \lfloor \frac{M}{2} \rfloor \}$, $\pi$'s $\rho$-robustness is the minimum of the average episode reward over all environments at radius $\rho$: 
\begin{gather} 
    \rho\text{-robustness} =  \min_{(i,j) \in \mathcal{G}_\rho} \mathbb{E}_{\tau \sim \pi, P_{i,j}} \left[ \sum^T_{t=1} R(s_t,a_t) \right] ,
\end{gather}
where $\mathcal{G_\rho} = \{ (i,j):||(i,j)-(\lfloor\frac{M}{2} \rfloor,\lfloor \frac{M}{2} \rfloor)||_\infty = \rho \}$ is the set of cells with $L_\infty$-norm distance $\rho$ to the grid center. 

\subsection{Experiments on Classic Control Tasks}

\begin{table*}[t]
\centering
\adjustbox{max width=0.98\linewidth}{%
\small
\begin{tabular}{c c c c c}
\toprule
\textbf{Method} & \textbf{PPO} & \textbf{PPO-GBR} & \textbf{PPO-PGD} & \textbf{PPO-PGDLC (Ours)} \\
\midrule
\diagbox[height=0.4cm]{$\rho$}{$\epsilon$} & -- &
\begin{tabular}{@{}ccccc@{}}
0.001 \ \ & 0.003 \ \ & 0.005 \ \  & 0.007 \ \ & 0.01
\end{tabular} &
\begin{tabular}{@{}ccccc@{}}
0.001 \ \ & 0.003 \ \ & 0.005 \ \ & 0.007 \ \ & 0.01
\end{tabular} &
\begin{tabular}{@{}ccccc@{}}
0.001 \ \ & 0.003 \ \ & 0.005 \ \ & 0.007 \ \ & 0.01
\end{tabular} \\
\midrule
0 & 498.46 &
\begin{tabular}{@{}ccccc@{}}
485.16 & \textbf{499.11} & 498.90 & 497.64 & 417.98
\end{tabular} &
\begin{tabular}{@{}ccccc@{}}
454.38 & 498.17 & 498.86 & 497.91 & 495.90
\end{tabular} &
\begin{tabular}{@{}ccccc@{}}
497.85 & 498.11 & 487.68 & 498.37 & 497.03
\end{tabular} \\
1 & 498.53 &
\begin{tabular}{@{}ccccc@{}}
485.93 & \textbf{499.03} & 491.70 & 496.90 & 368.73
\end{tabular} &
\begin{tabular}{@{}ccccc@{}}
398.28 & 498.03 & 496.04 & 497.36 & 493.61
\end{tabular} &
\begin{tabular}{@{}ccccc@{}}
497.67 & 497.95 & 479.79 & 498.02 & 494.21
\end{tabular} \\
2 & 498.35 &
\begin{tabular}{@{}ccccc@{}}
458.64 & \textbf{498.35} & 484.74 & 491.13 & 362.82
\end{tabular} &
\begin{tabular}{@{}ccccc@{}}
372.98 & 497.91 & 496.20 & 495.98 & 491.81
\end{tabular} &
\begin{tabular}{@{}ccccc@{}}
496.75 & 497.85 & 473.41 & 492.85 & 485.27
\end{tabular} \\
3 & 469.98 &
\begin{tabular}{@{}ccccc@{}}
312.67 & 389.71 & 484.37 & 368.36 & 367.31
\end{tabular} &
\begin{tabular}{@{}ccccc@{}}
366.20 & 425.62 & 461.67 & 428.17 & 412.21
\end{tabular} &
\begin{tabular}{@{}ccccc@{}}
401.61 & \textbf{494.07} & 476.45 & 486.14 & 391.46
\end{tabular} \\
4 & 252.30 &
\begin{tabular}{@{}ccccc@{}}
190.31 & 366.12 & 306.90 & 235.82 & 355.25
\end{tabular} &
\begin{tabular}{@{}ccccc@{}}
347.56 & 332.35 & 352.66 & 190.94 & 363.45
\end{tabular} &
\begin{tabular}{@{}ccccc@{}}
198.69 & 390.34 & 344.58 & \textbf{417.69} & 249.88
\end{tabular} \\
5 & 223.29 &
\begin{tabular}{@{}ccccc@{}}
179.05 & 240.70 & 306.05 & 247.43 & 212.51
\end{tabular} &
\begin{tabular}{@{}ccccc@{}}
161.07 & 333.17 & 318.14 & 196.87 & 320.19
\end{tabular} &
\begin{tabular}{@{}ccccc@{}}
214.75 & \textbf{357.11} & 142.79 & 313.83 & 250.34
\end{tabular} \\
\bottomrule
\end{tabular}
}
\adjustbox{max width=0.98\linewidth}{%
\begin{tabular}{c c c c c}
\toprule
\textbf{Method} & \textbf{PPO} & \textbf{PPO-GBR} & \textbf{PPO-PGD} & \textbf{PPO-PGDLC (Ours)} \\
\midrule
\diagbox[height=0.4cm]{$\rho$}{$\epsilon$} & -- &
\begin{tabular}{@{}ccccc@{}}
0.001 \quad & 0.003 \quad & 0.005 \quad & 0.007 \quad & 0.01
\end{tabular} &
\begin{tabular}{@{}ccccc@{}}
0.001  \quad & 0.003 \quad  & 0.005  \quad & 0.007  \quad & 0.01
\end{tabular} &
\begin{tabular}{@{}ccccc@{}}
0.001  \quad & 0.003  \quad & 0.005  \quad & 0.007  \quad & 0.01
\end{tabular} \\
\midrule
0 & 6339.91 &
\begin{tabular}{@{}ccccc@{}}
5553.56 & 5700.36 & 6112.06 & 1666.71 & 174.35
\end{tabular} &
\begin{tabular}{@{}ccccc@{}}
6346.54 & 5997.84 & 5696.34 & 468.64 & 180.53
\end{tabular} &
\begin{tabular}{@{}ccccc@{}}
\textbf{7010.88} & 6985.20 & 5822.36 & 6941.87 & 3076.43
\end{tabular} \\
1 & 6220.70 &
\begin{tabular}{@{}ccccc@{}}
5800.68 & 5200.72 & 5655.35 & 1438.12 & 171.33
\end{tabular} &
\begin{tabular}{@{}ccccc@{}}
5667.91 & 5754.71 & 5373.74 & 417.02 & 165.94
\end{tabular} &
\begin{tabular}{@{}ccccc@{}}
6646.53 & \textbf{6885.73} & 5387.60 & 6548.76 & 2694.21
\end{tabular} \\
2 & 5536.80 &
\begin{tabular}{@{}ccccc@{}}
5263.08 & 4760.70 & 5052.55 & 1438.12 & 165.99
\end{tabular} &
\begin{tabular}{@{}ccccc@{}}
5366.59 & 5399.01 & 4828.06 & 352.18 & 149.04
\end{tabular} &
\begin{tabular}{@{}ccccc@{}}
5996.85 & \textbf{6262.51} & 5118.59 & 5725.31 & 2541.31
\end{tabular} \\
3 & 4993.21 &
\begin{tabular}{@{}ccccc@{}}
4722.58 & 4235.75 & 4430.63 & 1472.58 & 163.42
\end{tabular} &
\begin{tabular}{@{}ccccc@{}}
4824.69 & 4714.49 & 4261.62 & 294.23 & 143.33
\end{tabular} &
\begin{tabular}{@{}ccccc@{}}
5210.44 & \textbf{5539.27} & 4545.94 & 4965.86 & 2460.67
\end{tabular} \\
4 & 4421.72 &
\begin{tabular}{@{}ccccc@{}}
4227.00 & 3796.47 & 3914.00 & 1439.15 & 166.53
\end{tabular} &
\begin{tabular}{@{}ccccc@{}}
4334.81 & 4211.22 & 3920.84 & 275.26 & 145.93
\end{tabular} &
\begin{tabular}{@{}ccccc@{}}
4519.73 & \textbf{4794.98} & 4081.62 & 4365.67 & 2239.75
\end{tabular} \\
5 & 4004.55 &
\begin{tabular}{@{}ccccc@{}}
3804.88 & 3419.03 & 3544.69 & 1396.46 & 175.12
\end{tabular} &
\begin{tabular}{@{}ccccc@{}}
3904.32 & 3763.67 & 3497.29 & 265.72 & 152.83
\end{tabular} &
\begin{tabular}{@{}ccccc@{}}
4040.66 & \textbf{4223.23} & 3667.65 & 3898.59 & 2123.52
\end{tabular} \\
\bottomrule
\end{tabular}
}
\caption{$\rho$-robustness metrics for all methods on Cartpole (top) and Ant (bottom); higher values indicate better robustness. 
Each entry reports the average episode reward over perturbed environments at radius $\rho$ for different uncertainty bounds $\epsilon$. The \textbf{bold} number represents the highest reward given $\rho$. Overall, PPO-PGDLC achieves the highest $\rho$-robustness across most tested radii and $\epsilon$ values, demonstrating stronger resilience to transition-dynamics perturbations compared with all baseline algorithms.}
\label{table:rho_robustness_diff_dynamics}
\end{table*}

\subsubsection{Baseline Algorithms} We compare PPO-PGDLC to three baseline algorithms: the standard  RL baseline PPO, and two robust RL variants, PPO-GBR and PPO-PGD. PPO-GBR employs the gradient-based regularizer described earlier. Both PPO-GBR and PPO-PGD set the regularization weight ($\lambda = 0$) so that no Lipschitz regularization is applied to the critic. PPO-PGD serves as an ablated version of PPO-PGDLC that retains the PGD-based worst-case value estimation but omits the Lipschitz-regularized critic. 

\subsubsection{Training Details} Both actor and critic networks are multilayer perceptrons with hidden sizes $[256,256]$ and learning rates of $3\times10^{-4}$.  For PPO-GBR, PPO-PGD and PPO-PGDLC, the uncertainty-set radius is varied as $\epsilon\in\{0.001,0.003,0.005,0.007,0.01\}$. To solve WCVE, PPO-PGD and PPO-PGDLC employ 10 PGD steps with a step size of $\epsilon/10$. In PPO-PGDLC, the weight $\lambda$ of local regularization term is set to $0.001$ for both Cartpole and Ant. Policies are trained in the nominal environment for $1.2\times 10^6$ transitions on Cartpole and $3\times 10^7$ transitions on Ant. Each configuration is repeated 4 times with different random seeds, and the evaluation results are averaged.

\subsubsection{Performance and robustness results} Fig.~\ref{fig:cartpole-ant-results} shows the training rewards for all four algorithms. On Cartpole, all policies achieve comparable performance, while on Ant, PPO-PGDLC converges to a higher final reward than the other algorithms. Next, we compare policy robustness against dynamics perturbations. We rescale the mass and static friction of all rigid bodies with factors within $[0.2, 1.8]$, creating $11\times 11 = 121$ environments.  
These two parameters are chosen because they directly influence the robot’s inertial and contact properties. Heatmaps in Fig. \ref{fig:cartpole-ant-results} visualize the performance across all perturbed environments. On Cartpole, when mass scale is in $\{0.2,0.36,0.52\}$, PPO-PGD and PPO-PGDLC achieve higher rewards than PPO-GBR and PPO, and all algorithms show comparable rewards for other mass scales. On Ant, PPO-PGDLC achieves higher rewards when mass scale is in $[0.2, 1.0]$ and friction scale is in $[1.0, 1.8]$, compared to the baselines. The heatmaps of PPO-PGD and PPO-PGDLC demonstrate the benefit of learning a Lipschitz-regularized critic for the WCVE problem. Table \ref{table:rho_robustness_diff_dynamics} presents the value of $\rho\text{-robustness}$ for all algorithms and $\epsilon$ values. We summarize our findings as follows. Firstly, PPO-PGDLC achieves the highest $\rho$-robustness value in 3 out of 6 radii on Cartpole, and across all 6 radii on Ant. Compared to PPO, PPO-PGDLC achieves 65.55\% $(\epsilon=0.007,\rho=4)$ and 59.93\% $(\epsilon=0.003,\rho=5)$ higher $\rho$-robustness value on Cartpole, and 9.87\% higher $\rho$-robustness value on average for all radii values on Ant. Additionally, compared to PPO-GBR and PPO-PGD, PPO-PGDLC improves policy performance for all radii when $\epsilon \leq 0.005$, with overall significant improvements when $\epsilon\geq0.007$. Finally, comparing all four algorithms reveals a trade-off between robustness and performance. Training on overly large uncertainty sets (e.g. $\epsilon\in \{0.007,0.01\}$) still impairs policy performance, which is consistent with the conclusion in \cite{kuang2022learning}, but learning a Lipschitz-regularized critic helps mitigate this effect.

\subsection{Experiments on the Unitree Go2 Robot}


We compare PPO-PGDLC to PPO on a real-world quadrupedal locomotion task, in terms of policy smoothness and performance. To ensure that the experimental results reflect the algorithms' effectiveness, neither algorithm is fine-tuned or exposed to real-world dynamics during simulation and hardware deployment. Specific policy configurations are detailed in the Appendix.

\subsubsection{Simulation Setup} 
\label{sec:go2-simulation-setup}
Our simulation environment is based on \cite{wtw-go2}, which adapts \textit{Walk-These-Ways} \cite{margolis2022walktheseways} to learn quadrupedal gaits with a Unitree Go2 robot. The observation vectors $o_t \in \mathbb{R}^{70}$ include torso orientation ($\mathbb{R}^3$), current joint positions ($\mathbb{R}^{12}$), current joint velocities ($\mathbb{R}^{12}$), actions from the last two time steps ($\mathbb{R}^{24}$), periodic clock input for four feet ($\mathbb{R}^{4}$), and behavior-specific commands ($\mathbb{R}^{15}$). The action vectors $a_t \in \mathbb{R}^{12}$ represent the desired joint positions, which are converted to joint torques via PD control with gains $k_p=25$ and $k_d=0.6$. We use the reward functions from \cite{margolis2022walktheseways} to learn two gaits (trotting and bounding) while enforcing gait stability and smoothness, with the following ranges for command velocities $[v^\text{cmd}_x,v^\text{cmd}_y,\omega^\text{cmd}_{\text{yaw}}]$: $v^\text{cmd}_x \in [-1,1]\text{m/s}$ for forward velocity, $v^\text{cmd}_y \in [-0.6,0.6]\text{m/s}$ for lateral velocity, and $\omega^\text{cmd}_{\text{yaw}}\in [-1,1]\text{rad/s}$ for yaw rate. Both simulation and hardware experiments run at a control frequency of 50Hz.

\subsubsection{Policy Setup} We compare PPO against PPO-PGDLC with the PGD solver modified, because the observation space defined above includes velocity and gait commands, previous actions, and clock inputs. Since this information is internal to the agent and not subject to noise in the environment, it is not realistic for it to be adversarially perturbed. Therefore, we mask the corresponding dimensions of the state vector so that they are not altered by the PGD solver. For PPO-PGDLC policies, we select $\epsilon=0.001$ and perform a grid search for the hyperparameter $\lambda \in \{5\times10^{-5},10^{-4},10^{-3},10^{-2}\}$. At each time step $t$, the policy receives the 5 most recent observations $\{o_{t-4},\cdots,o_t\}$ to predict an action $a_t$. Both actor and critic networks are multi-layer perceptrons with hidden sizes $[512, 256, 128]$ and ReLU activations. Each policy is trained for $1.97 \times 10^8$ transitions in total.


\subsubsection{Estimated Local Lipschitz Constant} To validate the effectiveness of local Lipschitz regularization in PPO-PGDLC, we estimate the local Lipschitz constant (LLC) of trained policies using \textit{auto-LiRPA} \cite{xu2020automatic}. This library is based on automatic linear relaxation to compute the bounds of network output within bounded input perturbations. For each policy, we collect 250 state transitions with trotting at commands $[v^\text{cmd}_x,v^\text{cmd}_y,\omega^\text{cmd}_{\text{yaw}}]=[0.5,0,0]$, then calculate the LLC of the actor and critic networks with $L_\infty$-norm perturbation bound 0.001. The results are shown in Table \ref{tab:go2-policy-llc}, and we observe that PPO-PGDLC achieves 57.7\% $\sim$ 96.6\% lower critic LLC and 7.0\% $\sim$ 27.6\% lower actor LLC than PPO, as $\lambda$ increases. Applying Lipschitz regularization to critic also enforce the smoothness of actor network (action prediction), but it does not imply robustness against dynamics perturbation. 

\begin{table}[!htbp]
\centering
\renewcommand{\arraystretch}{1.1} 
\resizebox{0.95\columnwidth}{!}{
\begin{tabular}{cccc}
\toprule
\textbf{Method} & $\lambda$ & \textbf{Actor LLC} & \textbf{Critic LLC} \\
\midrule
\textbf{PPO} & -- & 56.728 & 5.303 \\  \hline
\addlinespace
\multirow{4}{*}{\shortstack{\textbf{PPO-PGDLC} \\ ($\epsilon=0.001$)}} 
& $10^{-2}$ & 41.071 (27.6\%) & 0.179 (96.6\%) \\  
& $10^{-3}$ & 47.712 (15.9\%) & 0.416 (92.2\%) \\  
& $10^{-4}$ & 49.520 (12.7\%) & 1.425 (73.1\%) \\  
& $5\times 10^{-5}$ &  52.761 (7.0\%) & 2.242 (57.7\%) \\
\bottomrule
\end{tabular}
}
\caption{Estimated LLC for the actor and critic networks during trotting with commands $[v^\text{cmd}_x,v^\text{cmd}_y,\omega^\text{cmd}_{\text{yaw}}]=[0.5,0,0]$. Parenthetical values show the relative decrease compared to the LLC of PPO actor/critic.
}
\label{tab:go2-policy-llc}
\end{table}

\subsubsection{Additional Evaluation Metrics} We introduce three additional metrics to evaluate policy smoothness and performance: (1) Action Smoothness (AS) for quantifying abrupt changes in consecutive actions, (2) Second-order Fluctuation Ratio (SFR) \cite{zhang2024robust} for evaluating action oscillations, and (3) Velocity Tracking Error (VTE) for measuring the velocity tracking on the ground plane. The definitions are as follows:
\begin{gather}
    \text{AS} = \mathbb{E}_{\tau \sim \pi} \left[ \frac{1}{T} \sum_{t=1}^{T} ||a_t - a_{t-1}||_1 \right], \\
    \text{SFR} = \mathbb{E}_{\tau \sim \pi} \left[ \frac{1}{T} \sum_{t=1}^T ||a_t - 2a_{t-1} + a_{t-2}||_1 \right], \\
    \text{VTE} =  \mathbb{E}_{\tau \sim \pi} \left[ \frac{1}{T} \sum_{t=1}^T ||v_{xy,t} - v_{xy,t}^\text{cmd}||_2 \right].
\end{gather}

\begin{table*}[t]
\centering
\resizebox{0.78\linewidth}{!}{
\begin{tabular}{c c c c}
\toprule
\multirow{2}{*}{\textbf{Method}} & \multirow{2}{*}{$\lambda$} & \textbf{Bounding} & \textbf{Trotting} \\
 & & \begin{tabular}{@{}cccc@{}}
$\rho=0$ \ & $\rho=1$ \ & $\rho=2$ \ & $\rho=3$
\end{tabular} & \begin{tabular}{@{}cccc@{}}
$\rho=0$ \ & $\rho=1$ \ & $\rho=2$ \ & $\rho=3$
\end{tabular}  \\
\midrule
\textbf{PPO} & -- & 
\begin{tabular}{@{}cccc@{}}
11.286 & 11.172 & 10.516 & 9.041
\end{tabular} & 
\begin{tabular}{@{}cccc@{}}
15.207 & 14.839 & 13.958 & \textbf{12.377}
\end{tabular} \\  \hline
\addlinespace
\multirow{4}{*}{\shortstack{\textbf{PPO-PGDLC} \\ ($\epsilon=0.001$)}}
& $10^{-2}$ & 
\begin{tabular}{@{}cccc@{}}
13.467 & 12.373 & 10.827 & \textbf{9.093}
\end{tabular} & 
\begin{tabular}{@{}cccc@{}}
15.660 & 15.045 & 13.671 & 11.740 
\end{tabular} \\
& $10^{-3}$ & 
\begin{tabular}{@{}cccc@{}}
\textbf{14.194} & \textbf{13.610} & \textbf{11.563} & 3.078
\end{tabular} & 
\begin{tabular}{@{}cccc@{}}
12.598 & 12.180 & 11.671 & 10.782
\end{tabular} \\
& $10^{-4}$ & 
\begin{tabular}{@{}cccc@{}}
\ 9.971 \ & \ 8.557 \ & \ 7.399 \ & 6.413
\end{tabular} & 
\begin{tabular}{@{}cccc@{}}
15.218 & 14.637 & 13.420 & 11.961  
\end{tabular} \\
& $5\times 10^{-5}$ & 
\begin{tabular}{@{}cccc@{}}
13.452 & 13.085 & 11.308 & 6.771
\end{tabular} & 
\begin{tabular}{@{}cccc@{}}
\textbf{15.817} & \textbf{15.620} & \textbf{14.352} & 11.153
\end{tabular}  \\
\bottomrule
\end{tabular}
}
\caption{$\rho$-robustness for Go2 policies (bounding and trotting with commands $[v^\text{cmd}_x,v^\text{cmd}_y,\omega^\text{cmd}_{\text{yaw}}]=[0.5,0,0]$. Each value represents the average episode reward over perturbed environments defined by variations in body mass and ground-contact friction; higher values indicate stronger robustness. 
Overall, PPO-PGDLC surpasses PPO across most radii, while the optimal regularization weight $\lambda$ differs by gait, with $\lambda = 10^{-3}$ performing best for bounding and $\lambda = 5\times10^{-5}$ for trotting, indicating that the most effective level of critic smoothing depends on the task dynamics.
}
\label{tab:go2-rho-combined-robustness}
\end{table*}
\begin{table*}[t]
\centering
\resizebox{0.78\textwidth}{!}{
\begin{tabular}{c c c c c}
\toprule
\multirow{2}{*}{\textbf{Weight}} & \multirow{2}{*}{$v_x^\text{cmd}(\text{m/s})$} & \textbf{AS} $\downarrow$ & \textbf{SFR} $\downarrow$ & \textbf{VTE} $\downarrow$ \\
 &  & \begin{tabular}{@{}cc@{}}
\textbf{PPO} & \textbf{PPO-PGDLC}
\end{tabular} & \begin{tabular}{@{}cc@{}}
\textbf{PPO} & \textbf{PPO-PGDLC}
\end{tabular} & \begin{tabular}{@{}cc@{}}
\textbf{PPO} & \textbf{PPO-PGDLC}
\end{tabular} \\
\midrule
\multirow{3}{*}{+2kg} & 0.6 & \begin{tabular}{@{}cc@{}}
4.501 \ \ \ \ \ \ \ &  \textbf{4.306} \ \ \ \ \ \ \
\end{tabular}  & \begin{tabular}{@{}cc@{}}
\textbf{3.725} \ \ \ \ \ \ \  & 3.767 \ \ \ \ \ \ \
\end{tabular}  &  \begin{tabular}{@{}cc@{}}
\textbf{0.242} \ \ \ \ \ \ \  & 0.258 \ \ \ \ \ \ \
\end{tabular} \\
& 0.8 &  
\begin{tabular}{@{}cc@{}}
4.757 \ \ \ \ \ \ \ &  \textbf{4.488} \ \ \ \ \ \ \
\end{tabular}  & \begin{tabular}{@{}cc@{}}
3.942 \ \ \ \ \ \ \  & \textbf{3.741} \ \ \ \ \ \ \
\end{tabular}  &  \begin{tabular}{@{}cc@{}}
0.354 \ \ \ \ \ \ \  & \textbf{0.306} \ \ \ \ \ \ \
\end{tabular} \\
& 1.0 &
\begin{tabular}{@{}cc@{}}
5.068 \ \ \ \ \ \ \ &  \textbf{4.685} \ \ \ \ \ \ \
\end{tabular}  & \begin{tabular}{@{}cc@{}}
4.252 \ \ \ \ \ \ \  & \textbf{3.812} \ \ \ \ \ \ \
\end{tabular}  &  \begin{tabular}{@{}cc@{}}
0.411 \ \ \ \ \ \ \  & \textbf{0.396} \ \ \ \ \ \ \
\end{tabular} \\ \hline
\addlinespace
\multirow{3}{*}{+4kg} & 0.6 & 
\begin{tabular}{@{}cc@{}}
4.841 \ \ \ \ \ \ \ &  \textbf{4.498} \ \ \ \ \ \ \
\end{tabular}  & \begin{tabular}{@{}cc@{}}
3.742 \ \ \ \ \ \ \  & \textbf{3.739} \ \ \ \ \ \ \
\end{tabular}  &  \begin{tabular}{@{}cc@{}}
\textbf{0.432} \ \ \ \ \ \ \  & 0.473 \ \ \ \ \ \ \
\end{tabular} \\
& 0.8 & 
\begin{tabular}{@{}cc@{}}
5.067 \ \ \ \ \ \ \ & \textbf{4.686} \ \ \ \ \ \ \
\end{tabular}  & \begin{tabular}{@{}cc@{}}
4.023 \ \ \ \ \ \ \  & \textbf{3.855} \ \ \ \ \ \ \
\end{tabular}  &  \begin{tabular}{@{}cc@{}}
\textbf{0.517} \ \ \ \ \ \ \  & 0.584 \ \ \ \ \ \ \
\end{tabular} \\
& 1.0 & 
\begin{tabular}{@{}cc@{}}
5.424 \ \ \ \ \ \ \ & \textbf{4.913} \ \ \ \ \ \ \
\end{tabular}  & \begin{tabular}{@{}cc@{}}
4.371 \ \ \ \ \ \ \  & \textbf{4.087} \ \ \ \ \ \ \
\end{tabular}  &  \begin{tabular}{@{}cc@{}}
0.668 \ \ \ \ \ \ \  & \textbf{0.526} \ \ \ \ \ \ \
\end{tabular} \\
\bottomrule
\end{tabular}
}
\caption{Comparison of PPO and PPO-PGDLC ($\lambda=5\times10^{-5}$) across different payloads and commands. Each entry reports AS, SFR, and VTE at commanded forward velocities $v_x^{\text{cmd}}\in[0.6,1.0]\,\text{m/s}$ with $v_y^{\text{cmd}}=0\,\text{m/s}$ and $\omega_{\text{yaw}}^{\text{cmd}}=0\,\text{rad/s}$. 
PPO-PGDLC exhibits consistently smoother actions (lower AS in all tests and lower SFR in 5 of 6 tests) and comparable or improved velocity tracking (lower VTE in half of the tests), demonstrating better transfer stability under payload variation.
}
\label{tab:go2-hardware-with-weights}
\end{table*}

\subsubsection{Policy Smoothness and Performance on Nominal Dynamics} 
\begin{figure}[t]
    \centering
    \includegraphics[width=0.95\linewidth]{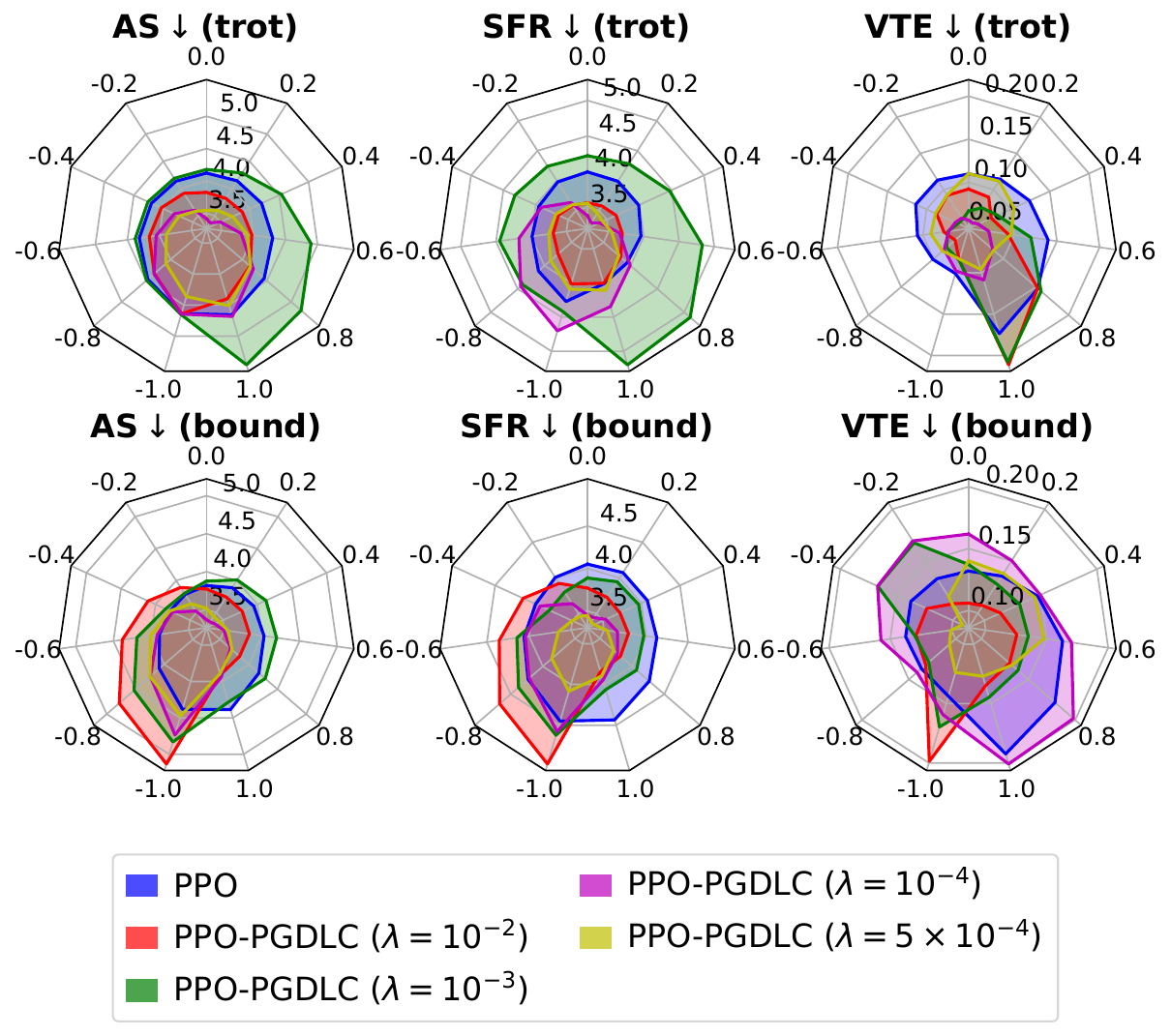}
    \alt{Radar chart visualizing Action Smoothness, Second-order Fluctuation Ratio and Velocity Tracking Error.}
    \caption{Radar charts visualizing AS, SFR and VTE across different command velocities. Each vertex represents a specific command velocity, with the distance from the center indicating the metric value. PPO-PGDLC with $\lambda = 5\times10^{-5}$ achieves the lowest AS, SFR and VTE for both gaits.}
    \label{fig:go2-radar}
\end{figure}

We firstly evaluate policy smoothness and performance for different command velocities and gaits on the nominal dynamics. For each policy, we collect 100 episodes for bounding and trotting gait with commands $v_x^{\text{cmd}} \in [-1,1] \, \text{m/s}$, $v^\text{cmd}_y=0\ \text{m/s}$ and $\omega^\text{cmd}_{\text{yaw}}=0\ \text{rad/s}$. The radar charts in Fig. \ref{fig:go2-radar} present the results for AS, SFR and VTE.  PPO-PGDLC with $\lambda=5\times 10^{-5}$ overall outperforms the other policies across the three metrics and both gaits, indicating superior policy smoothness and velocity tracking. However, $\lambda=5\times 10^{-5}$ is not necessarily optimal from a policy robustness perspective.

\subsubsection{Policy Robustness against Dynamics Perturbation} 
\label{sec:go2-policy-robustness-against-environmental-perturbation}
We next analyze policy robustness against dynamics perturbation via the $\rho$-robustness metric. A total of $7\times 7=49$ environments are created by rescaling body friction within the set $\{0.4, 0.6, 0.8, 1.0, 1.5, 2.0, 2.5\}$ and adding mass in $\{-4.5, -3.0, -1.5, 0.0, 1.5, 3.0, 4.5\}\text{kg}$ to the torso. For each environment, we collect 25 episodes for both bounding and trotting gait under the command velocities $[v^\text{cmd}_x,v^\text{cmd}_y,\omega^\text{cmd}_{\text{yaw}}]=[0.5,0,0]$. Table \ref{tab:go2-rho-combined-robustness} shows that PPO-PGDLC generally outperforms PPO (7 out of 8 radii), and the best-performing $\lambda$ differs by gait: $\lambda=10^{-3}$ for bounding versus $\lambda=5\times 10^{-5}$ for trotting. This demonstrates that the optimal regularization strength is task-specific.

\subsubsection{Zero-Shot Transfer to Hardware} 
Finally, we compare our trained policies in real-world settings without any fine-tuning. The Unitree Go2 robot, carrying additional payloads of 2\,kg and 4\,kg, is commanded to trot at different forward velocities. We measure AS, SFR, and VTE for command velocities $v_x^\text{cmd} \in [0.6,1.0] \, \text{m/s}$, where the real forward velocity of robot is estimated by a Kalman filter state estimator with low pass filter. Results in Table~\ref{tab:go2-hardware-with-weights} show that PPO-PGDLC achieves consistently smoother actions (lower AS in all tests and lower SFR in five of six cases) and comparable or improved velocity tracking (lower VTE in three of six cases), demonstrating better transfer stability under varying payloads. Fig.~\ref{fig:go2-hardware} in the Appendix shows the deployment of the PPO-PGDLC policy on hardware.

\begin{figure}[!htbp]
    \centering
    \begin{subfigure}{0.15\textwidth}
        \centering
        \includegraphics[width=\textwidth]{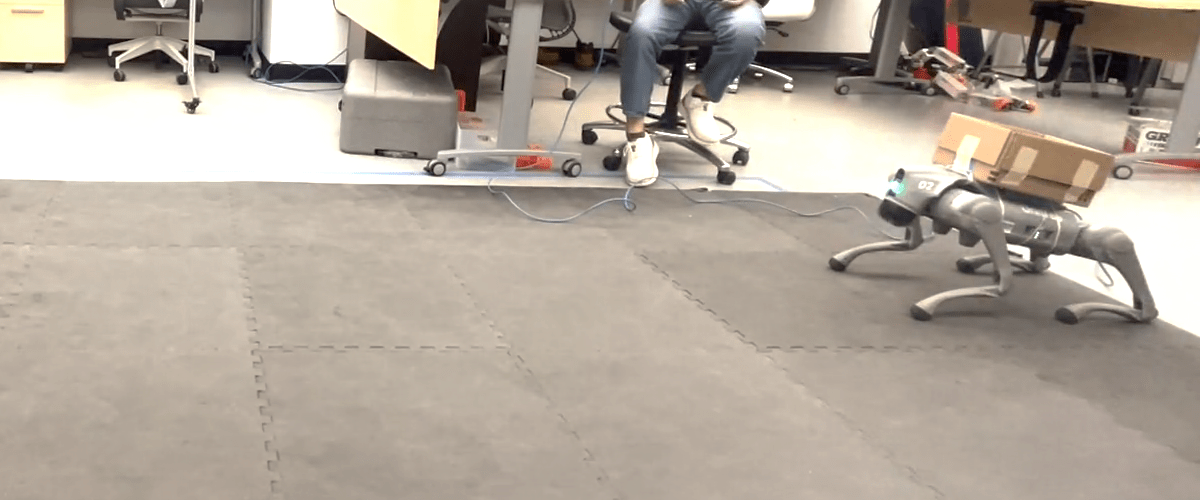}
    \end{subfigure}%
    \hfill
    \hfill
    \begin{subfigure}{0.15\textwidth}
        \centering
        \includegraphics[width=\textwidth]{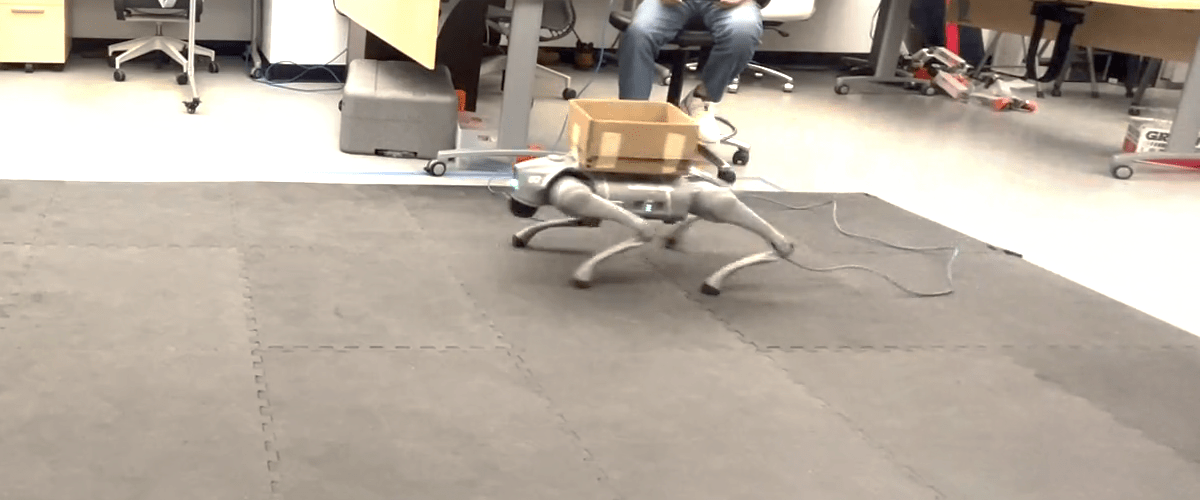}
    \end{subfigure}%
    \hfill
    \hfill
    \begin{subfigure}{0.15\textwidth}
        \centering
        \includegraphics[width=\textwidth]{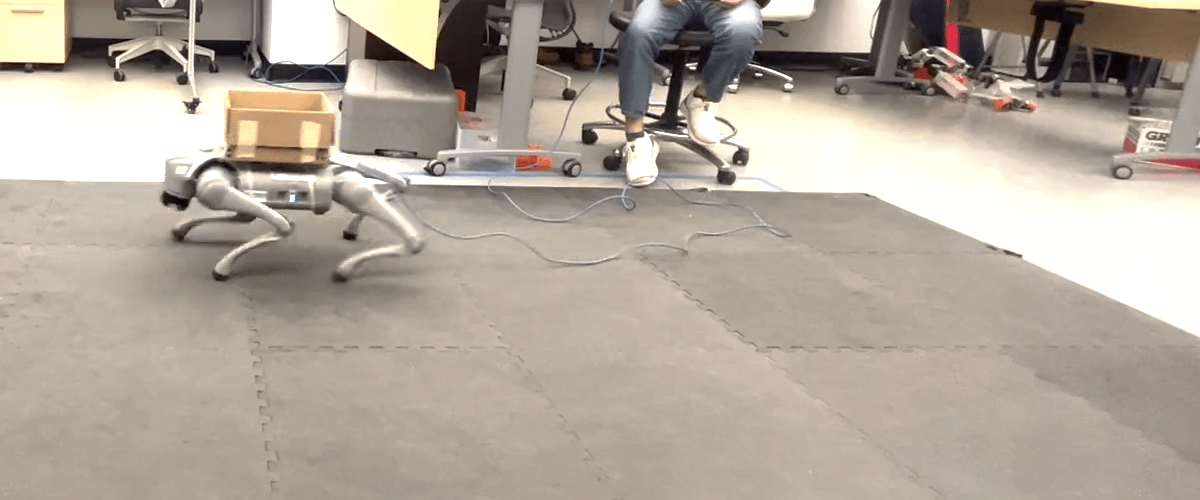}
    \end{subfigure}

    \smallskip
    
    \begin{subfigure}{0.15\textwidth}
        \centering
        \includegraphics[width=\textwidth]{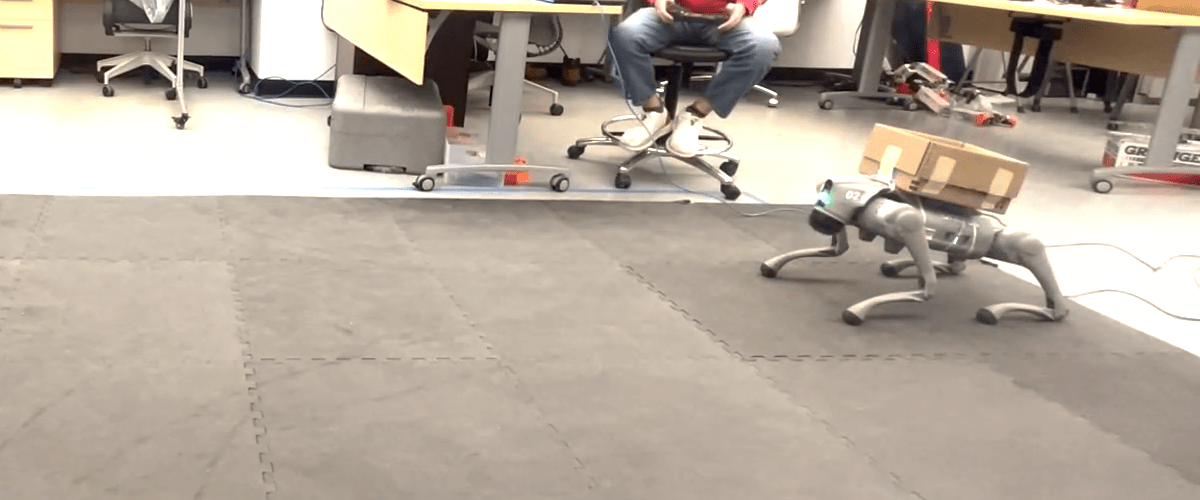}
    \end{subfigure}%
    \hfill
    \begin{subfigure}{0.15\textwidth}
        \centering
        \includegraphics[width=\textwidth]{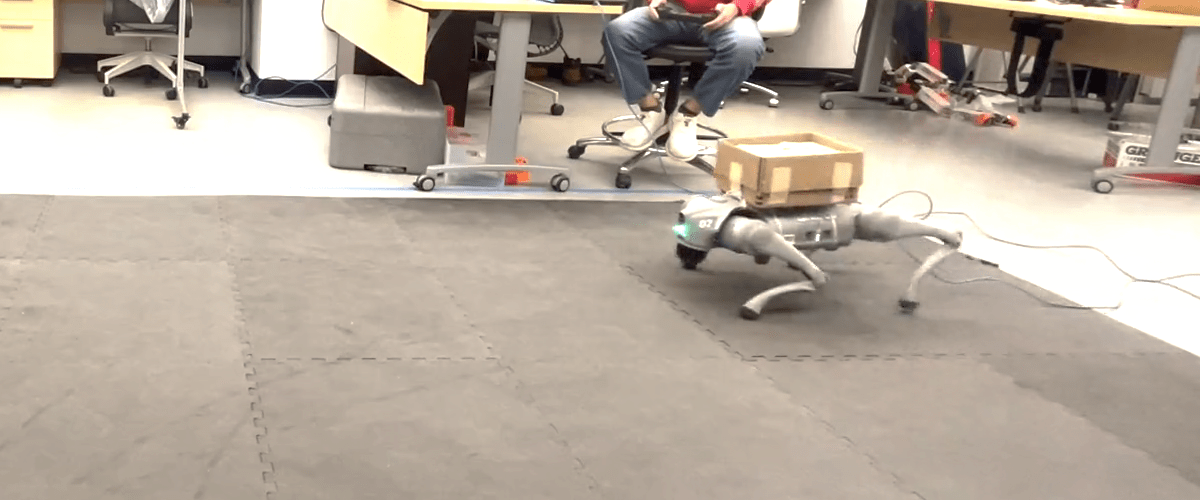}
    \end{subfigure}%
    \hfill
    \begin{subfigure}{0.15\textwidth}
        \centering
        \includegraphics[width=\textwidth]{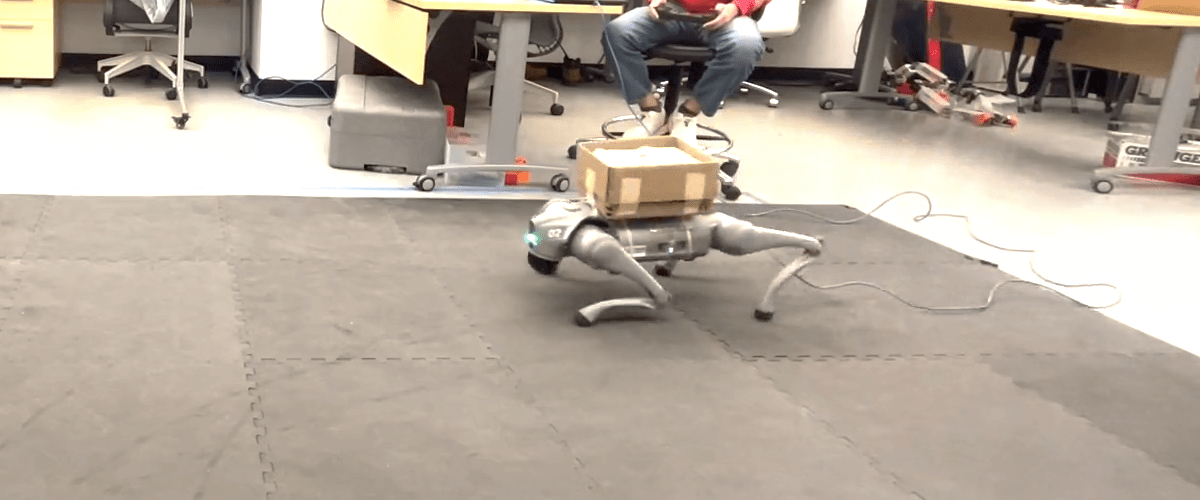}
    \end{subfigure}%
    \alt{Test PPO-PGDLC on Go2 hardware.}
    \caption{PPO-PGDLC policy on hardware, with a 2kg (top row) and 4kg (bottom row) payload and the command $v_x^\text{cmd}=1\text{m/s}$.}
    \label{fig:go2-hardware}
\end{figure}

\section{Conclusions}

In this paper, we introduce PPO-PGDLC, which leverages a Lipschitz regularized critic to improve policy smoothness and robustness under scenarios where the transition dynamics may be perturbed. The empirical results demonstrate the effectiveness of PPO-PGDLC for a range of simulated and real-world tasks. Future work could explore more general solutions for handling various transition dynamics and perturbations in the real world, such as distributionally robust MDP formulations.

\bibliographystyle{flairs}
\bibliography{ref}

\end{document}